\documentclass[letterpaper]{article} 
\usepackage{aaai24}  
\usepackage{times}  
\usepackage{helvet}  
\usepackage{courier}  
\usepackage[hyphens]{url}  
\usepackage{graphicx} 
\usepackage{natbib}  
\usepackage{caption} 
\usepackage{algorithm}
\usepackage[utf8]{inputenc} 
\usepackage[T1]{fontenc}    
\usepackage{microtype}
\usepackage{graphicx}  
\usepackage{booktabs}
\usepackage{hyperref}
\usepackage{algorithm}
\usepackage{algcompatible}
\usepackage{amsmath}
\usepackage{soul}
\usepackage[authormarkuptext=name,addedmarkup=bf,authormarkupposition=left]{changes}
\usepackage{xspace}
\usepackage{dsfont}
\usepackage{bm}
\usepackage{bbm}
\usepackage{nicefrac}       
\usepackage[scaled=.8]{beramono} 
\usepackage{amsthm}
\usepackage{amsfonts}
\usepackage{amssymb}
\usepackage{array}
\usepackage{mathtools}
\usepackage{caption}
\usepackage{subcaption}
\usepackage{fontawesome}
\usepackage{thmtools}
\usepackage{thm-restate}
\usepackage{lipsum}
\usepackage{pifont}
\usepackage{makecell}
\usepackage{tablefootnote}
\usepackage[tight-spacing=true]{siunitx}
\usepackage{multirow}
\usepackage{afterpage}
\usepackage{tabularx} 
\usepackage{url}
\usepackage{tikz}
\usepackage{indentfirst}
\usepackage{cleveref}
\usepackage{enumitem}
\usepackage{epsdice}
\usepackage{wrapfig}
\usepackage[export]{adjustbox}
\usepackage{xparse}

\makeatletter
\newcommand{\subalign}[1]{%
  \vcenter{%
    \Let@ \restore@math@cr \default@tag
    \baselineskip\fontdimen10 \scriptfont\tw@
    \advance\baselineskip\fontdimen12 \scriptfont\tw@
    \lineskip\thr@@\fontdimen8 \scriptfont\thr@@
    \lineskiplimit\lineskip
    \ialign{\hfil$\m@th\scriptstyle##$&$\m@th\scriptstyle{}##$\hfil\crcr
      #1\crcr
    }%
  }%
}
\makeatother

\Crefname{figure}{Figure}{Figures}
\crefname{figure}{Figure}{Figures}
\crefname{table}{Table}{Tables}
\crefformat{table}{Table~#1}
\crefformat{equation}{equation (#1)}
\Crefformat{Equation}{Equation (#1)}
\crefformat{algorithm}{Algorithm~#1}

\hypersetup{
  colorlinks=true,
  citecolor=blue,
  linkcolor=blue,
  urlcolor=blue
}

\newcommand\sdots{\hbox to 1em{.\hss.\hss.}} 
\DeclareMathAlphabet\mathbfcal{OMS}{cmsy}{b}{n} 

\DeclareMathSymbol{\shortminus}{\mathbin}{AMSa}{"39}

\setitemize[0]{leftmargin=*}

\newtheorem{definition}{Definition}

\newif\ifcomments
\commentstrue

\ifcomments
	\newcommand{\mr}[1]{\color{orange}MR: (#1)\color{black}\xspace}  
	\newcommand{\ml}[1]{\color{red}ML: (#1)\color{black}\xspace}  
	\newcommand{\me}[1]{\color{red}ME: (#1)\color{black}\xspace}  
	\newcommand{\dk}[1]{\color{red}DK: (#1)\color{black}\xspace}  
	\newcommand{\jf}[1]{\color{red}JF: (#1)\color{black}\xspace}  
	\newcommand{\jh}[1]{\color{red}JH: (#1)\color{black}\xspace}  
	\newcommand{\gt}[1]{\color{red}GT: (#1)\color{black}\xspace}  
\else
    \newcommand{\mr}[1]{}  
	\newcommand{\ml}[1]{}  
	\newcommand{\me}[1]{}  
	\newcommand{\dk}[1]{}  
	\newcommand{\jf}[1]{}  
	\newcommand{\jh}[1]{}  
	\newcommand{\gt}[1]{}  
	\newcommand{\cc}[1]{}  
\fi
\usepackage{newfloat}
\usepackage{listings}
\DeclareCaptionStyle{ruled}{labelfont=normalfont,labelsep=colon,strut=off} 
\floatstyle{ruled}
\newfloat{listing}{tb}{lst}{}
\floatname{listing}{Listing}

\newtheorem{example}{Example}[section]
\title{A Neuro-Symbolic Approach to Multi-Agent RL for Interpretability\\and Probabilistic Decision Making}
\author {
    Chitra Subramanian\textsuperscript{\rm 1},
    Miao Liu\textsuperscript{\rm 1},
    Naweed Khan\textsuperscript{\rm 2},
    Jonathan Lenchner\textsuperscript{\rm 1},
    Aporva Amarnath\textsuperscript{\rm 1},
    Sarathkrishna Swaminathan\textsuperscript{\rm 3},
    Ryan Riegel\textsuperscript{\rm 4},
    Alexander Gray\textsuperscript{\rm 1}
}
\affiliations {
    \textsuperscript{\rm 1}IBM T.J. Watson Research Center\\
    \textsuperscript{\rm 2}IBM Research Africa\\
    \textsuperscript{\rm 3}IBM Almaden Research Center\\
    \textsuperscript{\rm 4}IBM Research\\
    cksubram@us.ibm.comm, miao.liu1@ibm.com, Naweed.Khan@ibm.com,\\ lenchner@us.ibm.com,  aporva.amarnath@ibm.com, Sarath.Swaminathan@ibm.com,\\ ryan.riegel@ibm.com, alexander.gray@ibm.com
}

\begin{document}

\maketitle

\begin{abstract}
Multi-agent reinforcement learning (MARL) is well-suited for runtime decision-making in optimizing the performance of systems where multiple agents coexist and compete for shared resources. However, applying common deep learning-based MARL solutions to real-world problems suffers from issues of interpretability, sample efficiency, partial observability, etc. To address these challenges, we present an event-driven formulation, where decision-making is handled by distributed co-operative MARL agents using neuro-symbolic methods. The recently introduced neuro-symbolic Logical Neural Networks (LNN) framework serves as a function approximator for the RL, to train a rules-based policy that is both logical and interpretable by construction. To enable decision-making under uncertainty and partial observability, we developed a novel probabilistic neuro-symbolic framework, Probabilistic Logical Neural Networks (PLNN), which combines the capabilities of logical reasoning with probabilistic graphical models. In PLNN, the upward/downward inference strategy, inherited from LNN, is coupled with belief bounds by setting the activation function for the logical operator associated with each neural network node to a probability-respecting generalization of the Fréchet inequalities. These PLNN nodes form the unifying element that combines probabilistic logic and Bayes Nets, permitting inference for variables with unobserved states. We demonstrate our contributions by addressing key MARL challenges for power sharing in a system-on-chip application.
\end{abstract}

\section{Introduction}
Runtime resource management, such as in computer Operating Systems (OS), has traditionally relied on heuristics-based rules to achieve performance targets. However, as systems evolve and scale to more complex, distributed and heterogeneous processing elements, these rules are no longer sufficient for ensuring optimal runtime efficiency. Multi-Agent Reinforcement Learning (MARL), using deep learning methods, is increasingly applied to such complex resource management problems for providing distributed control that can dynamically adapt to changes in workload conditions and resource availability. However, although MARL deep learning methods can be powerful tools, there are three key challenges relating to real world resource management applications: (i) interpretability of policy solutions that permits human interactive control to ensure system reliability and resilience; (ii) sample efficiency during training  for known hard optimization problems requiring hefty combinatorial search; (iii) predictive control under noisy conditions, uncertainty and partial observability.   Solving these challenges requires a paradigm shift since current Deep Neural Network (DNN) models produce black box policies that lack human interpretability in the decisions, require extensive training, cannot extend much beyond the training set, and the large model sizes result in slow response times for runtime control. This calls for a system that supports a combination of neural-style learning with symbolic logical rules-based reasoning capability, where domain knowledge can be incorporated to reduce the search space during training, and Human-AI interaction is enabled in implementation through an interpretable policy. We propose neuro-symbolic AI techniques built upon a distributed multi-agent reinforcement learning framework to find tractable solutions for adaptive runtime resource management challenges.

\begin{figure}[h]
\captionsetup[subfigure]{skip=0pt, aboveskip=0pt}
\centering
\begin{subfigure}[b]{0.45\textwidth}
   \includegraphics[width=1\linewidth]{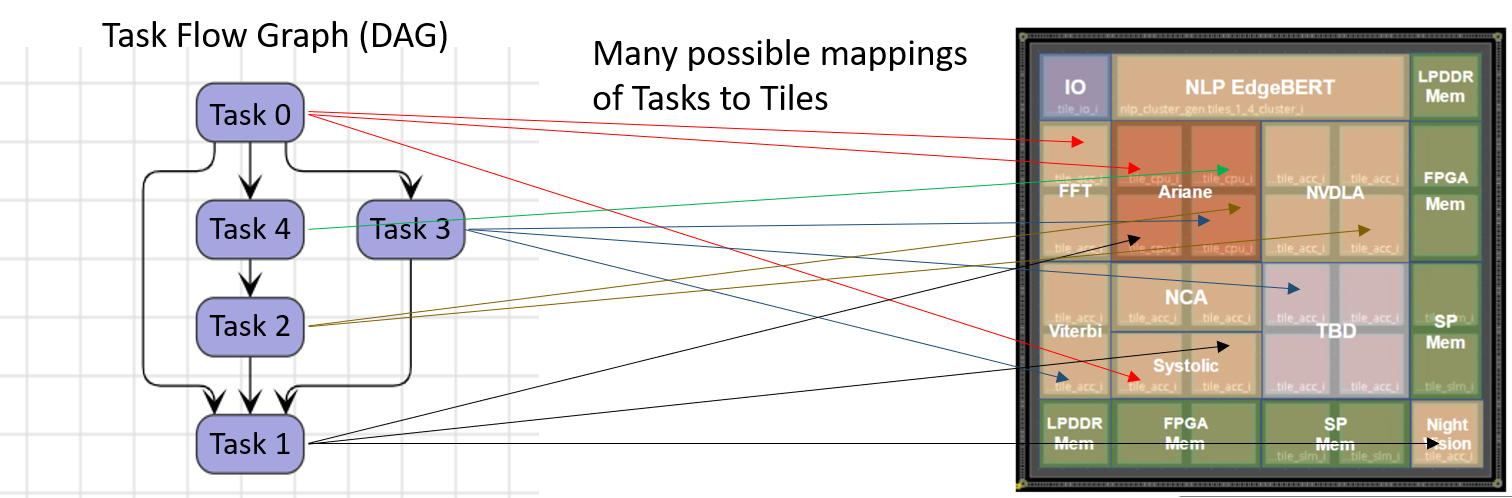}
        \caption{}
   \label{fig:martop_soc}
\end{subfigure}
\vskip 0.1in
\begin{subfigure}[b]{0.45\textwidth}
   \includegraphics[width=1\linewidth]{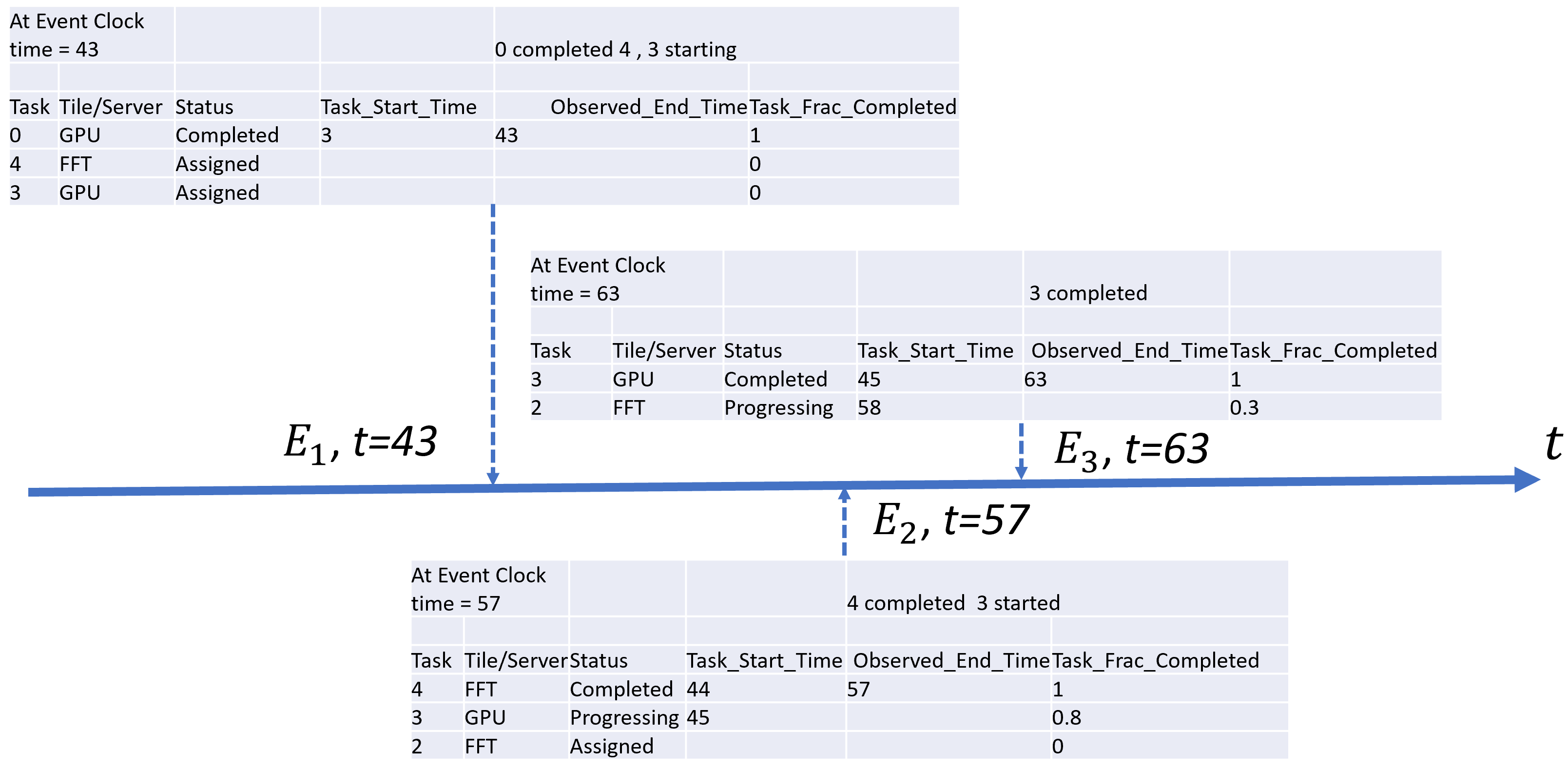}
   \caption{}
   \label{fig:martop_event}
\end{subfigure}
    \vskip-0.0in
    \captionsetup{font=small}
    \caption{
    \textbf{(a)} A job DAG capturing the dependencies of tasks and a mapping of tasks to tiles; \textbf{(b)} An example of runtime information of the job.}
    \label{fig:martop_example}
\end{figure}

We illustrate our neuro-symbolic approach to runtime resource management using MARL with a Heterogeneous System-on-Chip (HSoC) application. As shown in Figure~\ref{fig:martop_soc}, the HSoC is made up of several processing elements (PE). Compute workloads to be executed on the HSoC are defined as dataflow graphs (DAG) connecting individual tasks that are each assigned to one specific PE for execution based on availability, by a dynamic runtime scheduler. At any given time, multiple tasks belonging to one or more workflow DAGs are expected to run in parallel and the HSoC has a maximum allowed power envelope that all active PEs can together consume. Since the power allocated to an active PE determines the completion time of the task executing on it, multiple PEs need to share the limited power resource of the HSoC optimally to maximize the workload throughput of the system. We assign one RL agent to each PE in our system to optimally manage the power needs of its task during execution so that the associated DAG completes in the shortest possible time. This requires optimization of the dynamic power sharing between multiple agents based on workload conditions of the system. We developed a novel approach for achieving this goal using neuro-symbolic methods for interpretable rule learning and probabilistic inference under uncertainty.

We use the recently introduced Logical Neural Networks (LNN) \cite{riegel2020logical} in conjunction with Inductive Logic Programs (ILP) to train MARL policies \cite{DBLP:conf/aaai/SenCRG22}.  We use first-order LNN predicates and RL training in a gym environment to generate interpretable, rules-based resource sharing policies. We develop a probabilistic LNN framework to predict latent system states under uncertainty, arising both from the inherent partial observability in MARL settings as well as from stochastic behavior of the processes that drive state variables. The real time inference prediction from PLNN is used to modify implemented rules at runtime for a dynamic optimization solution.

\section{Preliminaries}
\subsection{Multiagent POMDPs}
Cooperation between multiple agents under uncertainty and partial observability can be represented by multiagent partially observable Markov decision processes (MPOMDPs)~\cite{NIPS2011_e5e63da7}, when their observations can be shared through communication. Specifically, an $n$-agent MPOMDP is defined as a tuple $\mathcal{M}_n\!=\!\langle I, S, O, A, P, Q, R, \gamma \rangle$;
$I\!=\!\{1,\sdots,n\}$ is the set of $n$ agents, 
$S$ is the set of states, 
$A$ and $O$ are respectively the cross-product of action and observation space for each agent, i.e. $A\!=\!\times_{i \in I} A^{i}$ and $O\!=\!\times_{i \in I} O^{i}$, $P\!:\!S\times{A}\mapsto S$ is the state transition function, $Q\!:\!S\times{A}\mapsto O$ is the set of observation function;
$R\!=\!\times_{i \in I} R^{i}$ is the set of reward functions, and
$\gamma\!\in\![0,1)$ is the discount factor.
Each agent $i$ executes an action at each timestep $t$ according to its stochastic policy $a^i_t\!\sim\!\pi^{i}(a^{i}_{t}|b(s_{t}),\theta^i)$ parameterized by $\theta^{i}$, where $s_{t}\!\in\!S$, and $b(s_t)$ is the belief state, which can be updated as 
\begin{eqnarray}
	\label{eq:beliefstate}
	&&\hspace{-0.28in}b(s_t) := p(s_t|b(s_0), {a}_0, O_1,...,{a}_{t-1}, O_t)
		\cr
&&\hspace{-0.27in}=\frac{p(O_t|{s}_t, {a}_{t-1})\sum_{s_{t-1}}p(s_t|s_{t-1}, a_{t-1})b(s_{t-1})}{\sum_{s_{t}}\sum_{{s}_{t-1}}p(O_t|s_t, {a}_{t-1})p(s_t|s_{t-1}, {a}_{t-1})b(s_{t-1})}.
\end{eqnarray}
A joint action ${a}_{{t}}\!=\!\{a^{i}_t,{a}^{{-i}}_{{t}}\}$ yields a transition from the current state $s_{t}$ to the next state $s_{t+1}\!\in\!{S}$ with probability ${P}(s_{t+1}|s_t,{a}_{{t}})$ and an observation $o_{t+1}\!\in\!{O}$ with probability ${P}(O_{t+1}|{s}_{t+1},{a}_{{t}})$, where the notation ${-i}$ indicates all other agents with the exception of agent $i$.
Agent $i$ then obtains a reward according to its reward function $r^{i}_t\!=\!R^i(s_t,{a}_{{t}})$. The goal is to optimize the global utility of the entire system measured by the accumulated total reward:
$R=\sum_{t=1}^H\sum_{i=1}^n\gamma^tr_t^i$. 
At the end of an episode, the agents collect a trajectory $\tau_{{\theta}}$ under the joint policy with parameters $\theta$, where $\tau_{\theta}\!:=\!(o_{0},{a}_{0},{r}_{0},\sdots,r_{H})$,  $\theta\!=\!\{\theta^{i},{\theta}^{-i}\}$ represents the joint parameters of all policies, ${r}_{{t}}\!=\!\{r^{i}_{t},{r}^{{-i}}_{{t}}\}$ is the joint reward, and $H$ is the episode horizon. 

\subsection{MARL for HSoC based on Event-driven MPOMDPs}
The classical MPOMDP framework is based on fixed time-step, where a decision is made at every time step. For the HSoC application considered in this work, power tokens are requested only when an event corresponding to the completion of an active task occurs. (This is illustrated by the example in figure~\ref{fig:martop_example}). Because agents do not need to make new decisions until an event is triggered, we formulate our problem as an event-driven MPOMDP. There are two major considerations in an event-driven process, namely 1) an environmental observation is obtained only when an event is triggered; 2) agents may receive rewards at any time steps, not just when an event prompt to act.

Before proceeding, we provide formal definition of events, which will be used throughout this paper:
\begin{definition}~\cite{de2014decision}
Let $F$ be a countable set. $F$ is a set of events over state space $S$ if there is a surjective mapping $\Phi:{S}\times{S}\rightarrow{{F}}$ such that, for all pairs of state transitions $<u,u'>$, $<v,v'>\in{S}\times{S}$:
\begin{itemize}
    \item if $\Phi(u,u') = \Phi(v,v')$, then $T(u,a,u') = T(v,a,v')$ and $p(\tau|a,u,u')=p(\tau|a, v,v') \forall a\in A$(events abstract transitions with the same stochastic properties); 
    \item $\Phi(u,u')\neq\Phi(u,u{''})$ if $u'\neq u^{''}$ (events univocally identify transitions from each state)
\end{itemize}
\end{definition}
Given the above definition, we now introduce Event-Driven Multiagent POMDP (ED-MPOMDP)~\cite{de2014decision}. Formally, ED-MPOMDP is a tuple $\mathcal{M}_n\!=\!\langle I, S, O, A, F, T, Y, C, \rho, \gamma \rangle$, where $I, S, O, A$ are the same as defined in MPOMDP;  $F$ is a set of events over $S$; $T\!:\!{S}\times A\mapsto F$ is the transition function, such that $T(s,{a},f)=p(f|s,{a})$ for $f\in{F}$, $\bm{a}\in A$, which represents the probability that $e$ will trigger while ${a}$ is being executed in s, thereby changing the state to $s'\in S$ such that $\Phi(s,s')=e$, and starting a new decision step; $Y\!:\!{F}\times{A}\mapsto O$ is the set of observation function, such that $Y(a,f,o)$ for $o\in O$, $f\in F$, $a\in A$; $C\!:\!S\times A\mapsto R^n$ is the set of instantaneous reward and $\rho\!:\!\times A\mapsto R^n$ is the set of reward rate. The final reward function is $R^i(s,a_i) =\rho^i(s,a)+\mathbb{E}_\tau[\int_0^\tau e^{-\gamma t}C^i(s,a)dt], \forall i\in I$.



The goal of HSoC power management is to maximize throughput (equivalent to minimize makespan), given random arrivals and completion deadline of jobs (described by DAGs). A node in a DAG (job) represents a task and has to be assigned to a processing element (PE), such as CPU, GPU, etc. Assuming there are finite number of PEs, each PE has an agent. Essentially, the problem is to develop a strategy for power sharing among sibling tasks at every cycle (clock time), so that the total time for completing all the tasks within the DAG is minimized. The agent-environment interaction is shown in Figure~\ref{fig:rlimplement} and consists of the following elements: 

\noindent\textbf{Agents:} Given tasks assigned, each agent calculates the number of power tokens needed within the current DAG in isolation ignoring other DAGs in the system.

\noindent\textbf{State space (}$S$\textbf{):} Each state $s\in S$ is a vector of state attributes. The state space $S=S_{\text{agent}}\cup S_{\text{environment}}$ can be decomposed into two subspaces:
(i) the agent-state $s_a\in S_\text{agent}$ consisting of information of each agent. For a task agent $i$ ($PE_i$), it knows 1) the maximum tokens allowed, 2) sibling tasks in a job (DAG), 3) PEs processing sibling tasks, 4) Sibling task computation time (w.r.t tokens requested), 
5) percentage of completion of each sibling task at event times, 6) ideal token/task time ratio between siblings; (ii) the environment-state $s_e\in S_\text{env}$ consisting of 1) High performance mode, 2) Plenty of Tokens, 3) Light Load, 4) Other DAGs' types (compute intense, memory intense), 5) High priority DAG, 6) Low Congestion and 7) Early Completion. These environment-state variables take binary values and are partially observable.

\noindent\textbf{Action space (}$A$\textbf{):} $A_i=\{a|a\in [0, 1, 2, ..., A_{max}]\}, \forall i\in I$, where $A_{max}$ is maximum power tokens allowed and it is pre-determined based on the DAG priority information.


\noindent\textbf{Reward function (}$R$\textbf{):} 
In an event-driven process, agents may receive rewards at any time steps, not just when an event prompt to act. To encourage agents to finish their tasks, the instantaneous reward is defined as an indicator function $\rho^i(s,a) = \mathbb{I}(\text{agent $i$ finishes an active task})$, which equals to 1 when the event an active task in $PE_i$ is finished , and $0$ otherwise. To balance the power assignment among sibling tasks so that the latency can be reduced, we define a per-step reward $r_t^i$, the design of which is detailed in the Appendix. Hence,the final reward agent $i$ received at the $j$th event (with event time $T_j$) is
$R^i_j = \rho_{T_j} + \sum_{t=T_{j-1}+1}^{T_j} r_t^i$, where the second term can be considered as reward accumulated between two events. 

\noindent\textbf{Trajectory Data}
For agent i, the data collected after processing a particular job (DAG) is $D^i_o = \{a^i_j, o^i_j, R^i_j\}_{j=1}^M$, with $i\in\{1,...,N\}$, where $M$ is the total number of event, which is equivalent to the number of nodes in the job DAG.
Since a POMDP can be converted into a belief-state MDP~\cite{kaelbling1998planning}, for the ease of deriving the policy learning algorithm, we can use models learned from historical data to compute the corresponding belief state according to~Eqn\eqref{eq:beliefstate}, leading to a converted trajectory $D^i_b = \{a^i_j, b^i_j, R^i_j\}_{j=1}^M$.

\noindent\textbf{Policy objective}
We apply Monte Carlo variant of policy gradient algorithm to multiagent reinforcement learning settings with centralized training and decentralized execution paradigm. The agents collect samples of an episode using its current policy, and use it to update the policy parameter $\theta$ with the objective function $J(\theta) = \mathbb{E}_{\tau\sim p_\theta(\tau^b)}[\sum_{t=0}^\infty\gamma^tr_t|\pi_\theta].$

Since $p_\theta(\tau^b) = p_\theta(a_1, b_1,...,a_T,b_T ) = p(b_1) \prod_{t=1}^T\pi_\theta(a_t|b_t)p(b_{t+1}|b_t,a_t)$, according to policy gradient theorem~\cite{sutton1999policy}, we have
\begin{equation}
	\label{eq:pg}
	\hspace{-0.0in}\approx\frac{1}{K}\sum_{k=1}^K\Big[\sum_t\log\pi_\theta(a_{k,t}|b_{k,t})\big(\sum_t\gamma^tr_t)\big)\Big].
\end{equation}
The policy parameter $\theta$ is updated according to the gradient ascent update rule $\theta\leftarrow \theta+\alpha\nabla_\theta J(\theta)$.


\newcommand{\cP}{\mathcal{P}}
\newcommand{\Luk}{Łukasiewicz~}
\newcommand{\Ft}{Fréchet~}
\newcommand{\Gd}{Gödel~}

\section{Probabilistic Logic and PLNN}

\subsection{Related Work} 
Probabilistic Logical Neural Networks, or PLNNs, build upon ideas contained in several other probabilistic and real-valued logic formalisms: Bayesian Networks, Credal Networks and their more recent variant, Logical Credal Networks, and Logical Neural Networks. For a brief discussion of these formalisms, see Appendix \ref{app:plnn-related}.

\subsection{Formal Specification of a PLNN} \label{sec:spec}

A PLNN is a logical-probabilistic graphical model. Formally, we define it to be a 4-tuple, $\cP = (V, E, B, J)$. 

$V$ is a finite collection of vertices or nodes. The nodes are partitioned into two groups:
\begin{itemize}
\item \textit{Propositional Nodes} ($V_P$) that are associated with `primitive' assertions about the state of the world (e.g., ``It will rain tomorrow.''). 
\item \textit{Operational Nodes} ($V_O$) that are associated with the logical operations $\wedge, \vee, \neg, \rightarrow, \equiv$ (identity) and also the conditional operator $|$.
\end{itemize}

$E$ is a set of directed edges between pairs of nodes. An edge either points from a propositional mode to an operational node or from one operational node to another operational node.

$B$ is a set of \textit{belief bounds} associated with each node in $V$. 
To each $v \in V$ we assign lower and upper bounds, $l_v, u_v$, with $0 \leq l_v \leq u_v \leq 1$, describing the  system's belief about the range of possible probabilities the primitive proposition or more complex logical proposition associated with $v$ can take on.

$J$ is a set of \textit{relative correlation coefficients} that are associated with the operational nodes $\wedge, \vee$ and $\rightarrow$. These correlations coefficients are subtly different than the more standard (Pearson) correlation coefficient.
They are also specified using bounds, but where the bounds range in $[-1, 1]$. We describe these in more detail in the next section.

\subsection{Motivating Concepts of PLNN}
There are three motivating concepts of PLNN that successively build upon  one another: (i) the so-called \Ft Inequalities, 
(ii) two methods of inference based on these inequalities that we refer to as ``upward'' and ``downward'' inferencing, and (iii) a method of interpolating the \Ft Inequalities using the previously introduced $J$ parameter. 
Initially we present just a simple intuitive picture of ``upward'' and ``downward'' inferencing. At the end, in a fourth subsection, we circle back to give the complete picture.

\subsubsection{Fr\'{e}chet Inequalities} \label{F Inequalities}
\begin{figure}[ht]
    \centering
    \includegraphics[scale=0.35]{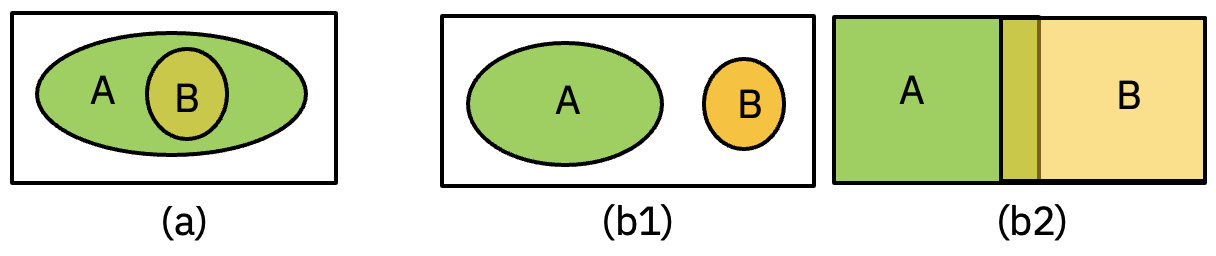}
    \captionsetup{font=small}
    \caption{The Venn Diagram associated with two propositions, $A$ and $B$, with fixed marginal probabilities $p(A) = p, p(B) = q$ in the case (a) where $A$ and $B$ are maximally  correlated and  the cases  (b1) and (b2) where $A$ and $B$ are maximally  anti-correlated.  
    }
    \label{fig:frechet_basic}
\end{figure}

Given two propositions $A$ and $B$, and some associated belief about the likelihood of the truth of each, that we denote by $p(A)$ and  $p(B)$ respectively, one cannot immediately say what one's belief should be about $A \vee B$. However, we can  say that for fixed values $p(A) = p$ and $p(B) = q$ that $p(A \vee B)$ is \textit{minimized} when $A$ and $B$ are maximally \textit{correlated} and, analogously, that $p(A \vee B)$ is \textit{maximized} when $A$ and $B$ are maximally \textit{anti-correlated}.  The propositions $A$ and $B$ are maximally correlated, whenever the less probable of the two is true, the other is true (Figure \ref{fig:frechet_basic}, case (a)). If they are equiprobable then one is true if and only if the other is true. As a result, we have that $p(A \vee  B) = \max(p(A), p(B))$. When $A$ and $B$ are maximally  anti-correlated, then, either $A$ and $B$ are never true together (Figure \ref{fig:frechet_basic}, case (b1)), or if their probabilities sum to more than $1$, they are true with as small a probability as possible (Figure \ref{fig:frechet_basic}, case (b2)).

Merging the two observations from the prior paragraph we get the two-sided inequality:
\small
\begin{equation}\label{eqn:2arg-frechet}
    \max(p(A), p(B)) \leq p(A \vee B) \leq \min(1, p(A) +p(B)).
\end{equation}
\normalsize
This inequality extends easily to the case of multiple propositional variables, $X_1,...,X_n$, to yield what is generally known as the Fr\'{e}chet Inequalities for disjunction \cite{frechet1934}:
\small
\begin{equation*}
    \max_i(p(X_i)) \leq p(X_1 \vee \cdot\cdot\cdot \vee X_n) \leq \min(1, \sum^n p(X_i)).
\end{equation*}
\normalsize

Moreover, with the addition of lower and upper bounds, $l_{X_i}, u_{X_i}$, for each of the propositional variables one easily obtains lower and upper bounds for the disjunction:
\begin{eqnarray}
 l_{X_1 \vee \cdot\cdot\cdot \vee X_n} &=& \max_i(l_{X_i}), \\
 u_{X_1 \vee \cdot\cdot\cdot \vee X_n} &=& \min(1, \sum_i u_{X_i}).
\end{eqnarray}

Analogous considerations lead one to the Fr\'{e}chet Inequalities for the conjunction of $n$ propositional variables:
\small
\begin{equation*}
    \max(0, \sum_{i = 1}^n p(X_i) - (n-1)) \leq p(X_1 \wedge \cdot\cdot\cdot \wedge X_n) \leq \min_i(p(X_i)),
\end{equation*}
\normalsize
The companion expressions for the lower and upper bounds of such a conjunction are then
\begin{eqnarray}
 l_{X_1 \wedge \cdot\cdot\cdot \wedge X_n} &=&  \max(0, \sum_{i = 1}^n l_{X_i} - (n-1)), \\
 u_{X_1 \wedge \cdot\cdot\cdot \wedge X_n} &=& \min_i(u_{X_i}).
\end{eqnarray}

These bounds further extend to implication, $p(A \rightarrow B)$ since implication is interpreted 
to be equivalent to $p(\neg A  \vee B)$. 

\subsubsection{Upward and Downward Inference} \label{sec:up-down-basic}

\begin{figure}
    \centering
    \includegraphics[scale=0.32]{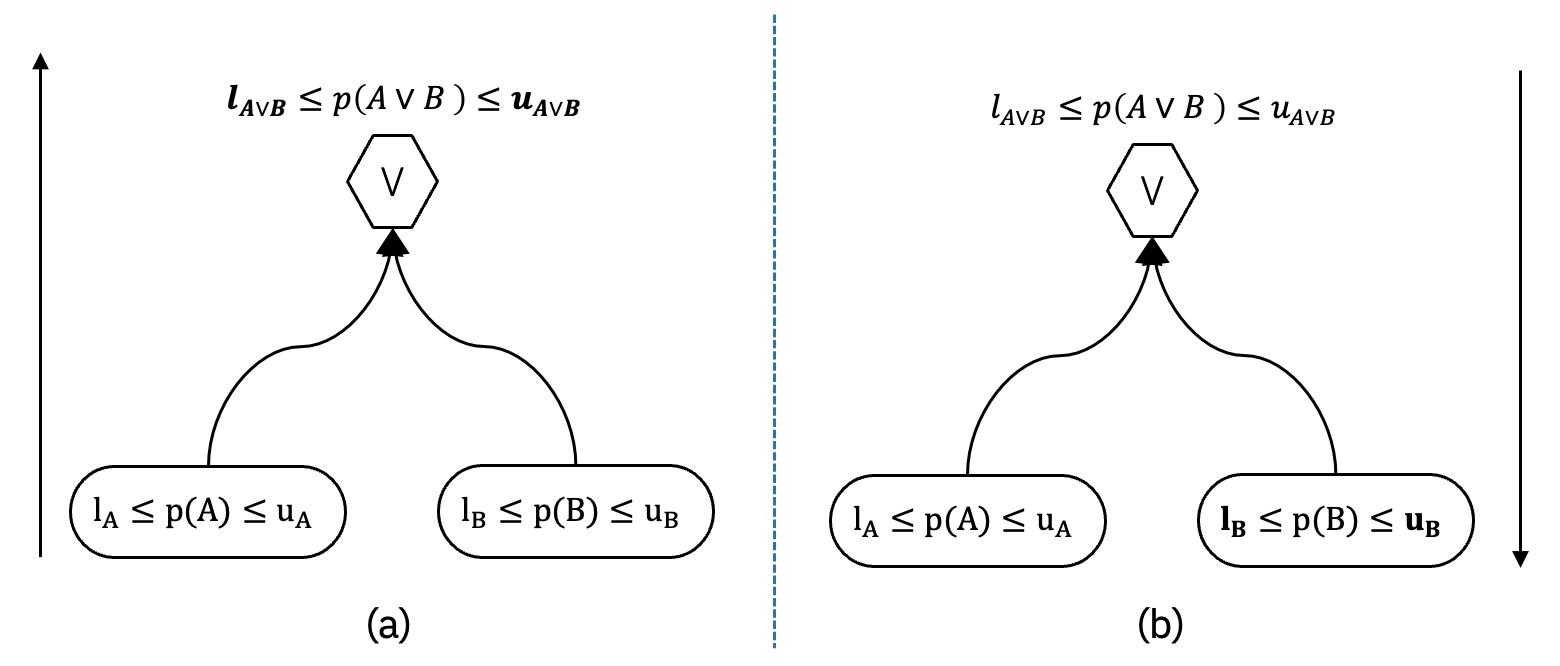}
    \captionsetup{font=small}
    \caption{(a) Simple upward inference in the case of operands, $A$ and $B$, coming into an OR operational node.  The bounds for the $\vee$ node are updated using the associated Fr\'{e}chet Inequalities (\ref{eqn:2arg-frechet}). (b) The analogous downward inference for the same set of nodes.  Using bounds on $A \vee B$ and $A$ we can back out inferred bounds on $B$.}
    \label{fig:simple_upward_and_downward_or}
\end{figure}

The next motivating concept of PLNN (borrowed from the earlier LNNs) is what we refer to as ``upward'' and ``downward'' inference.  We present a simplified version of the concept and then return to it later on. We refer to the process of updating bounds on an operational node, based on bounds on its inputs as ``upward inference''. This picture, with two input nodes coming into a disjunction is depicted in Figure \ref{fig:simple_upward_and_downward_or}a. The bounds are updating using the Fr\'{e}chet Inequalities for disjunction (\ref{eqn:2arg-frechet}).

Slightly less obviously, this process can be inverted. Sometimes we have  knowledge, say in the form of a prior, about the bounds on an operational node, that taken together with bounds on one or more of its input operands, allows us to update our  belief  bounds about a remaining operand. This  process, again in the case  of a two argument/operand disjunction is displayed in Figure \ref{fig:simple_upward_and_downward_or}b.
For example, in the figure, $A$ may represent the proposition ``It will next rain on a Monday,  Tuesday or Wednesday'', and $B$ may represent the proposition ``It will rain next on a Thursday, Friday, Saturday or Sunday''. We of course know for certain that $A \vee B$ is true (so that in this toy example we would have $l_{A \vee B} = u_{A  \vee B} = 1$). The rule for updating bounds on $B$ given known bounds on $A \vee B$ is simply:
\begin{eqnarray}
    l_B &\mapsto& \max(l_B, l_{A \vee B} - u_A), \label{downward-1}\\
    u_B &\mapsto& \min(u_B, u_{A \vee B}). \label{downward-2}
\end{eqnarray}
Downward inference for conjunction and implication works similarly.

\medskip

Upward and downward inference rules for conditional expressions of the form $(A|B)$ come directly from the definition of conditional probability and are given in Appendix \ref{app:up_down_condtional}. 

\smallskip

The basic computation mechanism in a PLNN is to perform successive iterations of upward and downward inferencing across all nodes in the PLNN graph until successive iterations fail to tighten any bounds by more than some $\epsilon > 0$. 

\subsubsection{$J$-Modulation of the \Ft Inequalities} \label{sec:j-mod}

The next motivating concept for PLNN is that it introduces a new parameter, which we call $J$, which for given marginal probabilities $p(A)$ and $p(B)$ interpolates between maximum anti-correlation ($J=-1$), statistical independence ($J=0$) and maximum correlation ($J=1$). $J$ is a mechanism for capturing prior knowledge of how the joint probability of $A$ and $B$ should behave as a function of the marginal probabilities even when the marginal probabilities are loose. That is, we track bounds on $J$ which may be tighter than the current bounds on the actual joint probability.

In the case of conjunction, the derivation of the $J$-modulated \Ft Inequalities of course begins with the \Ft Inequalities:
\begin{equation*}
    \max(0, p(A) + p(B) - 1) \leq p(A \wedge B) \leq \min(p(A), p(B)),
\end{equation*}
where the left hand side (the minimum value) corresponds to $J = -1$ and the right hand side (the maximum) corresponds to $J = 1$. At $J=0$ we want $p(A \wedge B) = p(A)p(B)$. We set:
\begin{eqnarray}
    P_{\wedge, -1} &=&  \max(0, p(A) + p(B) - 1), \label{p_and_-1}\\
    P_{\wedge, 0} &=& p(A)p(B), \label{p_and_0} \\
    P_{\wedge, 1} &=& \min(p(A),p(B)). \label{p_and_1}
\end{eqnarray}

We then obtain the $J$-modulated \Ft Inequalities for conjunction by setting:
\footnotesize
\footnotesize
\begin{equation} \label{moduldated-frechet-and}
    P_\wedge(J) = \frac{-J(1-J)}{2}P_{\wedge, -1} + (1+J)(1-J)P_{\wedge, 0}
                                                    + \frac{J(1+J)}{2}P_{\wedge, 1}.
\end{equation}
\normalsize

Note that we have specifically arranged that $P_\wedge(-1) = P_{\wedge, -1}, P_\wedge(0) = P_{\wedge, 0},$ and $P_\wedge(1) = P_{\wedge, 1}$.  The $J$-modulated \Ft Inequalities for disjunction and implication are derived similarly.

\medskip

%
%

\subsubsection{Upward and Downward Inference Revisited}

When we described upward and downward inference earlier in this section, we purposely oversimplified the discussion because we had not yet discussed interpolation of the \Ft Inequalities using the $J$ parameter. 
The more complete picture, including the $J$ parameter, is described in Appendix \ref{app:j-update}.

\section{Experimental Results and Discussion}

\subsection{LNN Rule Learning Implementation}
\begin{figure}[h]
\captionsetup[subfigure]{skip=0pt, aboveskip=0pt}
\centering
\begin{subfigure}[t]{0.32\textwidth}
		\centering
			\includegraphics[scale=0.45]{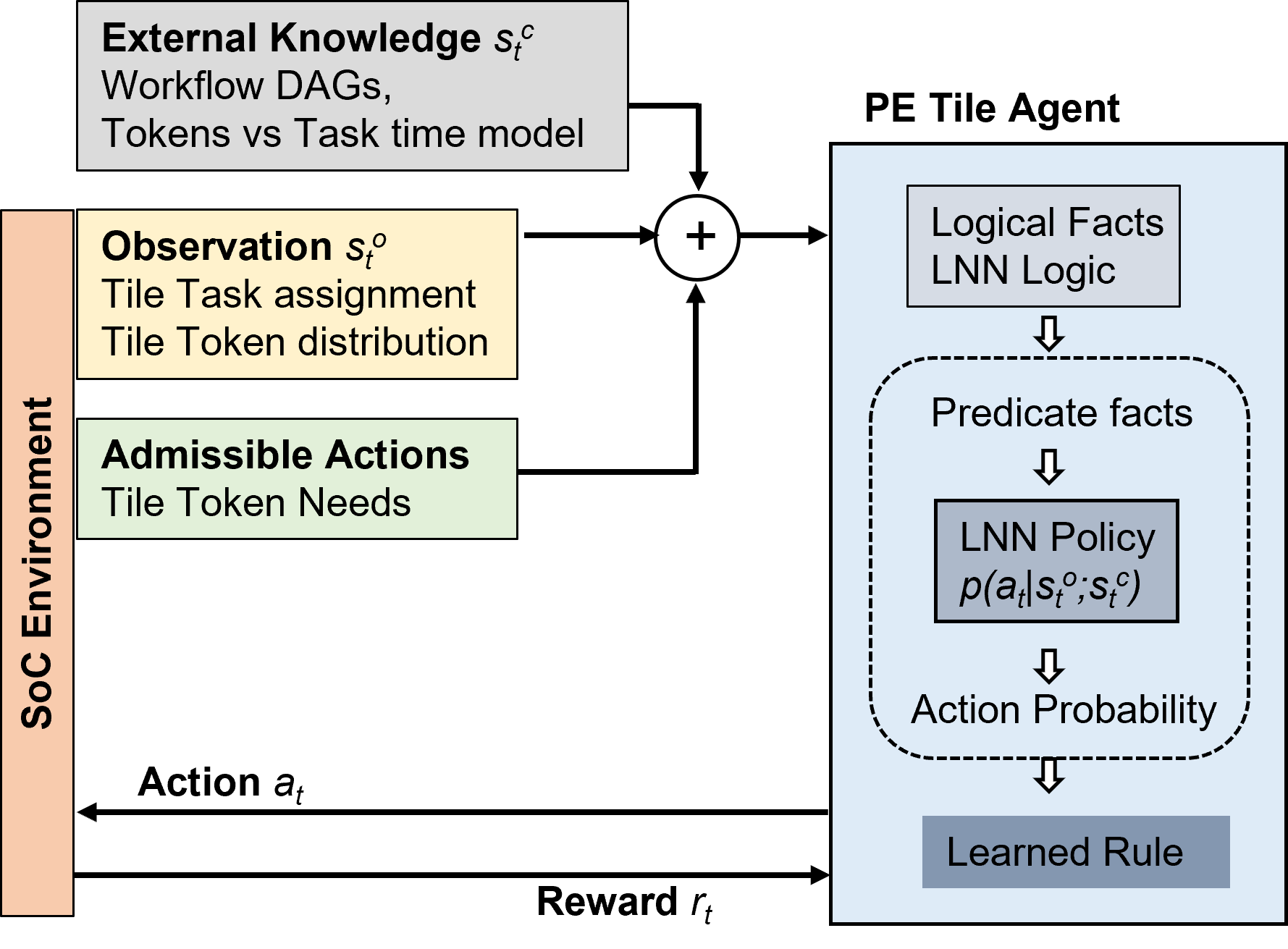}
		\caption{}\label{fig:rlimplement}
	\end{subfigure}
\vskip 0.1in
\begin{subfigure}[t]{0.32\textwidth}
		\centering
		\includegraphics[scale=0.45]{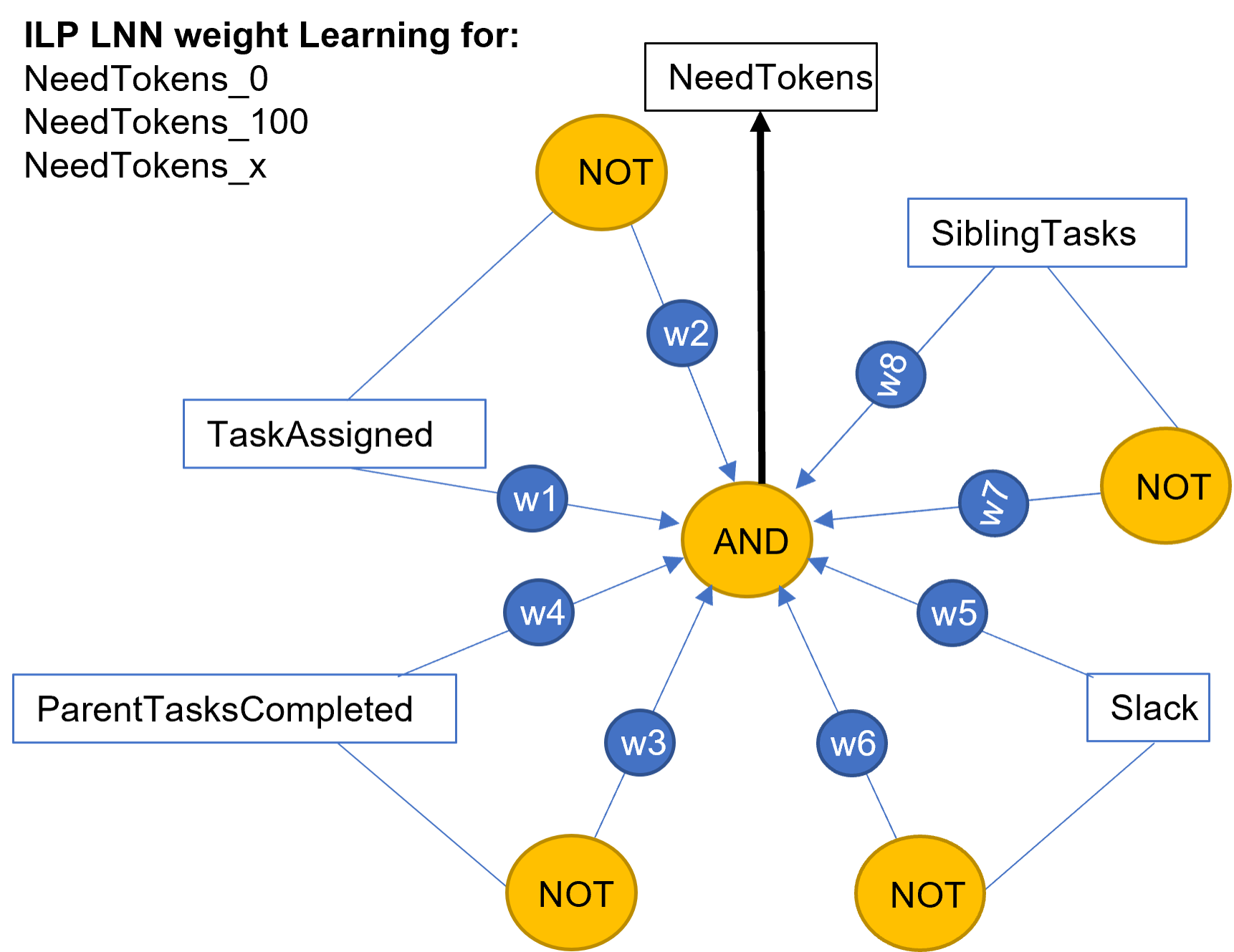}
    \caption{}
    \label{fig:lnntemplate}
	\end{subfigure}
    \vskip-0.0in
    \captionsetup{font=small}
    \caption{
    \textbf{(a)} A job DAG capturing the dependencies of tasks and a mapping of tasks to tiles; \textbf{(b)} An example of runtime information of the job.}
    \label{fig:martop_example}
\end{figure}


Figure~\ref{fig:rlimplement} shows the implementation of RL training in which LNN serves as the differentiable function approximator for static rule learning. Each training episode consists of completing a single job or workflow DAG. The SoC Env keeps track of the task assignment, progress of tasks, and power token distribution between tiles during the lifetime of a job. At each time step the PE Tile agent takes the action of allocating a discrete percentage of maximum allowed power tokens for itself. The model was trained using Reinforce algorithm with PyTorch BCE loss function and Adamax optimizer and a dense reward structure where the reward function was based on completion time of all concurrent tasks within the DAG. These aspects follow a standard RL policy training setup. The specific changes made for LNN RL training were: (i) use of first-order logical predicates and logical observations; (ii) incorporating domain expert knowledge; (iii) control of admissible actions to impose guard rails. Each of these aspects will be discussed in detail next.

(i) Examples of logical predicates include ones that define whether a task is assigned to the agent’s PE Tile (‘TaskAssigned’), whether any parallel tasks exist at that time step (‘SiblingTasks’), and whether parent tasks have completed (‘ParentTaskCompleted’). The conversion of raw observation data into first-order logical predicates creates a higher level of abstraction that both improves interpretability as well as reduces the dimensionality of the problem as compared to a traditional RL or a propositional logic description. 

(ii) Expert knowledge about the features of the agent’s assigned task, such as the task’s contribution to the DAG critical path timing and relative duration compared to other concurrent tasks, are provided during training through a real-valued predicate (‘Slack’) having values between 0 and 1. In the most general form, multiple ‘Slack’ predicates, which are convex mathematical functions of the token distribution between concurrent tasks, would be available to learn optimization for different metrics based on where their peaks occur. For example, in our current illustration, the optimizing Slack function peaks when the token distribution between concurrent tasks in the same DAG predicts that the task completion times are in the same ratio as their ideal target task sub-deadline times. The rationale for this optimization function is that it tries to offset the uneven acceleration between tasks resulting from scheduler’s PE assignments by adjusting the token distribution to compensate, resulting in the shortest possible DAG completion time for the assigned PE combination. Thus, the Slack predicate allows various domain heuristics based mathematical optimization functions to be easily incorporated into the LNN rule learning system. 

(iii) Guard rail rules take care of edge cases such as forbidding the agent from taking the action of allocating 0\% power to itself while processing a task or taking all 100\% power while a sibling task is active. These 3 aspects allow greater capacity for human interaction and control in the MARL system, reduce training time and produce fully interpretable rules. 

A Łukasiewicz fuzzy logic conjunction with weighted input predicates and negated predicates was defined as the template for LNN rule learning as shown in Figure~\ref{fig:lnntemplate}. Inductive Logic Programming (ILP) along with double-description method of LNN rule learning [Sen et al., 2022] was used to learn interpretable First Order Logic LNN rules.

\subsection{Phase1 Training Method}
The ability of LNN to generate compact, rules-based, light weight policies, is demonstrated in the Phase 1 Training Results section. This enables fast response times and the logic-based rules allow inference extension well beyond the training set, which we will illustrate with our inference results.

\noindent\textbf{Phase 1 Training Results:} 
\begin{figure}
    \centering
    \includegraphics[scale=0.60]{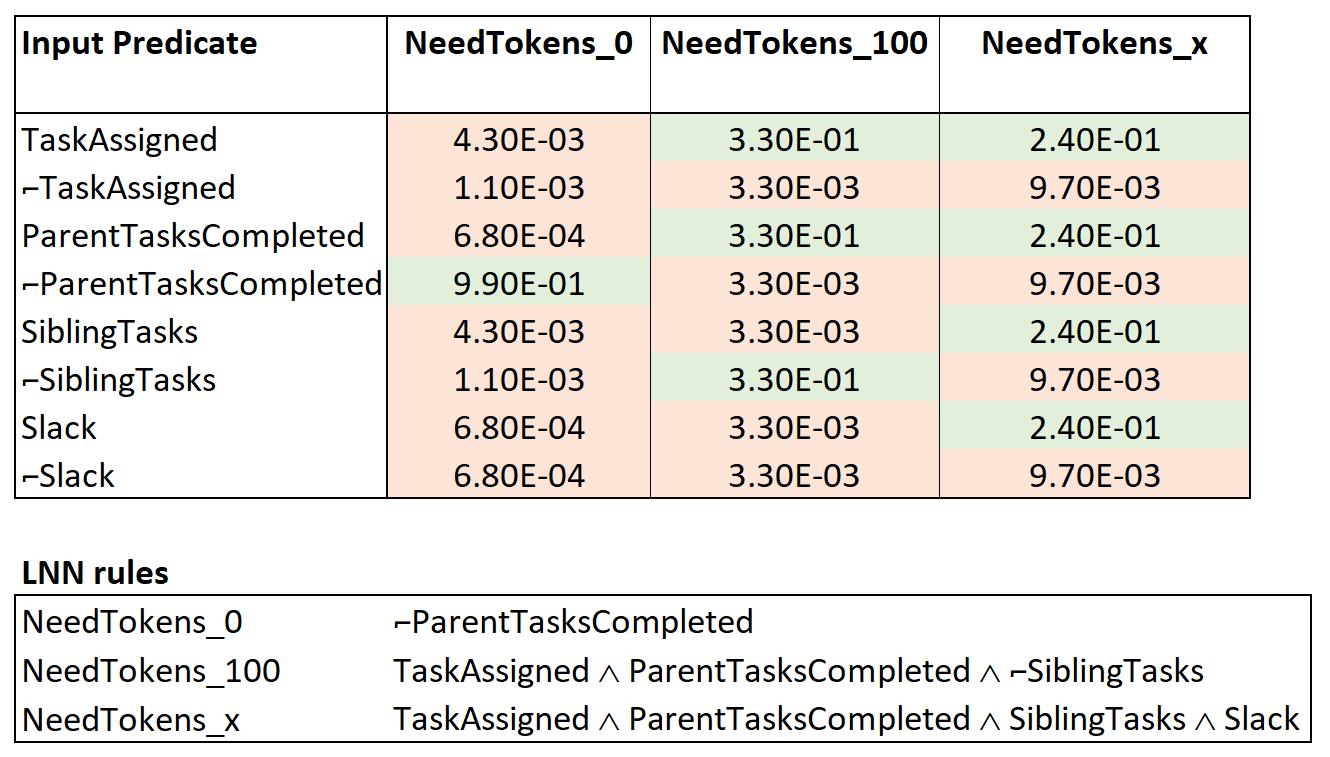}
    \captionsetup{font=small}
    \caption{Learned LNN weights for agent tokens requests and interpretable rules based on weight threshold of 0.1.}
    \vskip-0.1in
    \label{fig:lnnrules}
\end{figure}
Distinct LNN rules were learned simultaneously during training for 3 classes of actions that the agent could take in requesting its share of power tokens, i.e, NeedTokens\_0 for 0\% of power to the PE Tile, NeedTokens\_100 for 100\% of power share, and NeedTokens\_x for x\% of power share where 0 < x < 100. Inductive Logic Programming (ILP) along with double-description method of LNN rule learning [Sen et al., 2022] was used to learn interpretable First Order Logic LNN rules. The weights learned are shown in Figure~\ref{fig:lnnrules}. A weight threshold of 0.1 was used to filter out irrelevant predicates for each logical rule statement, resulting in the final rules as shown in Figure~\ref{fig:lnnrules}.
The learned rules are straight forward to interpret. The PE Tile agent requests 0\% power to the PE whenever its ‘ParentTasksCompleted’ predicate is NOT observed to be logically True. This can occur either because the PE is waiting for the parent tasks of its assigned task to complete, or it has not been assigned a task and is idle. Under both conditions the PE Tile is in standby mode and needs to be allocated 0\% power, hence the learned rule for NeedTokens\_0. On the other hand, when a task is assigned to a PE Tile and its parent tasks have completed, it needs to have power allocated for task execution. If there are no sibling tasks to run in parallel all available HSoC power needs to be allocated to the PE for minimizing the DAG makespan, hence the NeedToken\_100 rule. However, when sibling tasks exist, the NeedTokens\_x rule is activated, and the ‘Slack’ predicate’s peak value corresponds to optimal share of power for the PE Tile. As previously stated, the ‘Slack’ predicate takes on values based upon pre-processed expert information about the task’s features in the context of its DAG. These ‘Slack’ predicate values range between 0 and 1 and determine the probabilistic activation for different x\% power selections. Therefore, the ‘Slack’ predicate can be viewed as a family of activation curves for different x\% power allocations that the RL system can learn from, under different system conditions and optimization targets. 
We have thus successfully demonstrated in phase 1 training: the ability to learn human interpretable LNN rules-based MARL policy, incorporating domain expert knowledge, and guard rail control of admissible actions to minimize training time. 
Extensibility to unseen examples will be illustrated by the inference results next.

\noindent\textbf{Phase 1 Inference Results:} 
\begin{figure}[ht]
    \centering
    \includegraphics[scale=0.56]{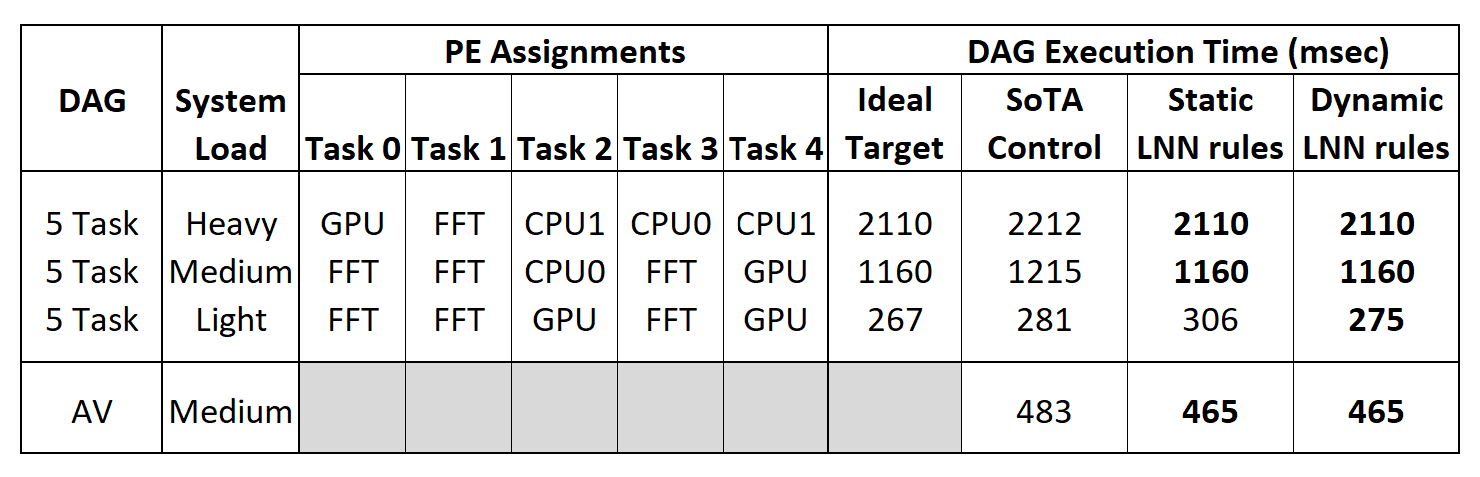}
    \captionsetup{font=small}
    \caption{Simulated DAG execution time for different test cases.}
    \label{fig:dagextime}
\end{figure}
The inference results provide evidence that using the learned LNN rules, PE agents are able to make token allocation decisions that improves DAG completion times as compared to the current state-of-the-art (SoTA) control case, where power tokens are shared equally between concurrent tasks at all times. Figure~\ref{fig:martop_example} shows the 5-task DAG used in Phase 1 training and Figure~\ref{fig:dagextime} shows the DAG execution time results for variable PE assignments under conditions of heavy, medium and light system loading. The Ideal Target DAG execution time is based on a fully static schedule of PE assignments, where all task assignments are known upfront for accurate token share optimization. This is not the reality of runtime scheduling where PE assignments are made dynamically just in time, depending on their availability. The static LNN rules-based power sharing cases outperform the control under heavy and medium system loading conditions, defined by the number of concurrent DAGs and availability of accelerator PEs, even for previously unseen PE assignments at training time. 

 To further demonstrate the extensibility of the LNN method, rules learned by training on the original 5-task DAG were used in executing a much more complex 42-task DAG for a real-world autonomous vehicle (AV) application and included in Appendix \ref{app:realdag}.  Once again, as shown in Figure~\ref{fig:dagextime}, the DAG completion time with LNN based power sharing  is better than the control. This illustrates the capabilities of neuro-symbolic LNN method, in simplifying MARL training and providing extensibility through logic. 
However, in the case of light system load, the SoTA control performed better than LNN rules. To understand this result we need to analyze the timing and uncertainty around the agent’s decisions. When a task is assigned at runtime, the agent is provided the PE assignments for future tasks in the DAG only for a limited time look ahead. Any power token decisions the agent may make to slow down its task at the start time could be completely negated if future unknown PE assignments place its siblings on much faster processors. In that case, the agent’s decision to slow down its task leads to a push out in the DAG’s completion time relative to a uniform power sharing control case. This situation is more probable if the system loading is light and high-speed accelerators are readily available for the scheduler to assign to future siblings. One solution would be for the agent to make a dynamic prediction of light system load and implement uniform power sharing to avoid the slowdown.  

\subsection{Dynamic Probabilistic Decision-Making}
While the rules generated in phase1 training offer full interpretability, they are static and do not respond to changes in system state variables, such as, system load conditions, which, based on domain knowledge, are expected to influence optimization decisions. For example, the stochasticity in workload intensity and arrival times results in unpredictable network congestion and hence, variation in task completion times. Additional noise sources include incomplete observations by agents, variability and offsets in the execution time model, etc. Therefore, along with the logical rules, there are several sources of noise in the system that need to be accounted for by applying marginal and conditional probabilities and correlations between predicates, in order to make accurate predictive decisions. PLNN concepts developed in this paper provide the framework for probabilistic inference about hidden states. In phase 2 of rule generation, additional domain knowledge is combined with offline statistical data and PLNN inferred predicate states, to construct dynamic rules for more optimal power sharing.

\subsubsection{PLNN Inference Experimental Results}

PLNN inference is used to enable dynamic predictions of a light system load, subject to partial observability of a subset of probabilistic bounds that are available at runtime. Analogously, this corresponds to performing inference on a CredalNet with only a subset of conditional-table rows being provided.
In Table \ref{tab:plnn_results}'s experiment, we use the domain knowledge from Appendix \ref{app:plnninfstp} to construct a PLNN graph, without providing additional logic for a full BayesNet treatment, e.g., conditional independencies. 
\begin{table}[ht]
    \begin{center}
    \scalebox{.70}{
\begin{tabular}{l|llll|}
\cline{2-5}
\multicolumn{1}{c|}{} & \multicolumn{1}{c}{Q1} & \multicolumn{1}{c}{Q2} & \multicolumn{1}{c}{Q3} & \multicolumn{1}{c|}{Q4} \\ \hline
\multicolumn{1}{|l|}{\textit{PLNN}} & (0.625, 0.667)* & (0.0, 0.286) & (0.625, 0.833)* & (0.0, 0.286) \\
\multicolumn{1}{|l|}{\textit{PLNN-J}} & (0.679, 0.681)* & (0.0, 0.286) & (0.625, 0.833)* & (0.0, 0.286)* \\ \hline
\multicolumn{1}{|l|}{Expected} & \multicolumn{1}{c}{High} & \multicolumn{1}{c}{Low} & \multicolumn{1}{c}{High} & \multicolumn{1}{c|}{Low} \\ \hline
\end{tabular}
    }
    \captionsetup{font=small}
    \caption{
    \textit{PLNN} correctly determines that the probability of a lightly loaded system lies within some threshold of the expected results. 
    The \textit{J}-modulated variant includes additional domain knowledge of expected correlations between variables from figure \ref{fig:plnn_correlations}, such that \(AC=(-1.0, 0.5), ID=0\text{, and }HC=(-0.5, 1.0)\).
    `*' indicates that a contradiction is present somewhere in the graph.}
    \label{tab:plnn_results}
    \end{center}
\end{table}

While the addition of the \textit{J} parameter did not provide any additional certainty about system loading, it significantly "tightened" the bounds for many other nodes in the graph (see figure \ref{fig:plnn_exp_viz}).

This visualisation further aids the discussion of an interpretable neuro-symbolic system, since PLNN inherits the per node computation of LNN. This allows the graph to be inspected and logs of the upward-downward algorithm to be inspected for a better understanding of each computational step taken towards the converged result.

We note that partial contradictions are present in the graph, this arises because each MARL agent aggregates information from multiple agents --- different agents could have different beliefs. This, however, should not be confused with the interpretation of a classical contradiction found in resolution-style theorem-proving. While in classical logic, an inconsistency proves that statements in the system violate the rules that are present, (P)LNN 
(i) identifies and arrests the point of contradiction, preventing contradictions from propagating; (ii) deduces the extent to which the violation occurs - allowing partial contradictions to occur at meaningful crossing points; (iii) and follows a deterministic inference procedure to help identify which statements were involved in the contradiction compute stack. 
This allows for the following key features: (i) the graph can be inspected to identify whether the contradiction is not relevant to the result, i.e., is it present within a subgraph that is not connected to an inspected nodes (i.e., `LightLoad'), or not involved in the computation thereof - select nodes in figure \ref{fig:plnn_exp_viz} fit this category; (ii) where a contradiction is present in a subgraph of interest, the computational log and original DAG can guide us to ``loosen'' flexible nodes, i.e., observations by an agent with short-term history, or observations derived from sibling agents further back in time; (iii) aggressive variable correlations can be relaxed, e.g., assertions of maximal-correlation can be reduced to not include as much support from anti-correlation terms. 

\subsubsection{Probabilistic Decision-Making Experimental Results}
Referring back to Figure~\ref{fig:dagextime}, the Dynamic LNN rules column uses PLNN predictive inference for system loading to dynamically modify the rules to use either LNN power sharing or the uniform power sharing. Whenever system light load is predicted by PLNN, uniform power sharing is used instead of the LNN power sharing rules. The results for all loading cases, including light load, show  results close to ideal target performance with this dynamic rule modification.  

\section{Conclusions}
 We have demonstrated LNN-based neuro-symbolic methods to address key issues encountered in applying MARL solutions to real world problems. We have shown that our approach produces interpretable logical rules, reduces the problem size, permits learning with fewer examples and extends the solution well beyond the training set to unseen cases. The novel PLNN method developed here, allows agents to make probabilistic prediction of the system state, despite limited partial observability, based on domain knowledge of correlation between variables and historical data for conditional probabilities. We used the example of a HSoC power sharing application to illustrate the advantages of applying neuro-symbolic techniques to practical MARL solutions.

\pagebreak
\bibliography{aaai24}

\appendix

\section*{APPENDICES}

\section{LNN Rule Learning Setup}
Each PE in the HSoC has a PE Tile agent that learns optimal power sharing rules based on variable DAGs and PE task assignments by an external scheduler. The PE power sharing allocation as a percent of maximum allowed HSoC power is referred to as ‘Tokens’. The maximum Token count in the system is, therefore, fixed at 100 and each PE Tile agent learns to request a share of the Tokens that would minimize the completion time of the DAG that is associated with its assigned task. The first phase of training assumes a deterministic task execution time model given the tokens for any PE. 

\begin{figure}[ht]
    \centering
    \includegraphics[scale=0.80]{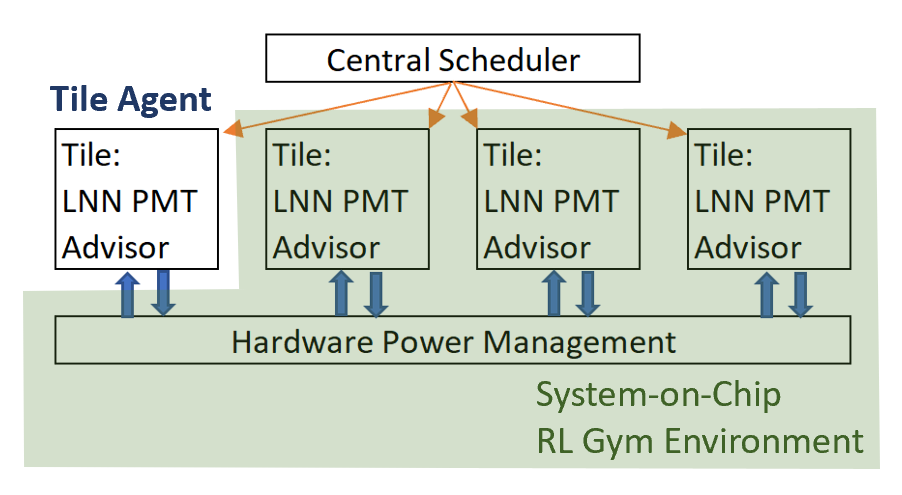}
    \caption{RL rule learning setup.}
    \label{fig:rlsetup}
\end{figure}

Figure~\ref{fig:rlsetup} 
shows this noise-free setup used for the first phase of training where the agent learns a set of static LNN rules using an RL Gym Environment. The PE Tile agent learns optimal PMT rules by performing the action of requesting a binned percentage of the maximum allowed HSoC power (Tokens) at each time step during RL training and receiving rewards based on the associated DAG completion time. In this centralized MARL training scenario, all agents have access to information regarding all other agents during training. Since the PE Tile agents operate independently and don’t interact with each other during the inference/operation phase, the training was done for one PE Tile agent at a time, considering all the other PE Tile agents to be part of the HSoC environment. In this setup, the use of first-order LNN input predicates, appropriately derived based on domain knowledge of the problem, as well as abstracting variations in individual PE tile response to power allocation by the use of a normalized uniform Tokens description, all the PE Tile agents ultimately learn essentially identical PMT rules. 

\section{Design of Per-step Reward Function}
To balance the power assignment among sibling tasks so that the latency can be reduced, we design a per step reward function of each agent as follows:
\[ r_{t}^i=
  \begin{cases}
    \sum_{j\in sibling(i)}0.25-0.25|\frac{h_i-h_j}{h_i+h_j}|       & ~~ \text{if } \text{agent $i$ has}\\& ~~ \text{sibling tasks;}\\
    cst*(fps-0.99)*0.25  & ~~ \text{otherwise.} \text{ }
  \end{cases}
\]
In the above equation, $h_i=tsd_x*fp_i$, is the headway made so far for the task $x$ assigned to $PE_i$ with $std_x$, the task subdeadline 
and $fp_i$ the calculated fractional progress per cycle for a token needs action selected by $PE_i$. Similarly, $h_j=tsd_y*fp_j$ is the headway $PE_j$ made so far for the task $y$ (the sibling task of $x$) assigned to $PE_j$ with $std_y$, the task subdeadline and $fp_j$ the calculated fractional progress per cycle for a selected token needs action by $PE_j$. 

When there are no siblings, the reward for agent $i$ is simply a function of $cst*fp_i$ with $cst$ being the current service time. $fp$ is calculated fractional progress per cycle for selected tokens. 

\section{Probabilistic Logic and PLNN Related Work} \label{app:plnn-related}

\subsection{Bayesian Networks} \label{sec:bayes-nets}
Graphical models such as Bayesian networks provide a powerful framework for reasoning about conditional dependency structures over many variables. A BN \cite{pearl88} is defined by a tuple $\langle \mathbf{X}, \mathbf{D}, \mathbf{p}, G \rangle$, where $\mathbf{X} = \{X_1, \ldots, X_n\}$ is a set of variables over multi-valued domains $\mathbf{D} = \{D_1, \ldots, D_n\}$, $G$ is a directed acyclic graph (DAG) over $\mathbf{X}$ as nodes, and $\mathbf{p} = \{p_i\}$ where $p_i = p(X_i|pa(X_i))$ are \emph{conditional probability distributions} (CPDs) associated with each variable $X_i$ and its parents $pa(X_i)$ in $G$. The network represents a joint probability over $\mathbf{X}$, namely:
\begin{equation}
p(X_1,\ldots,X_n) = \prod_{i=1}^{n} p(X_i|pa(X_i))    
\end{equation}



\begin{figure}[ht]
    \centering
    \includegraphics[scale=0.30]{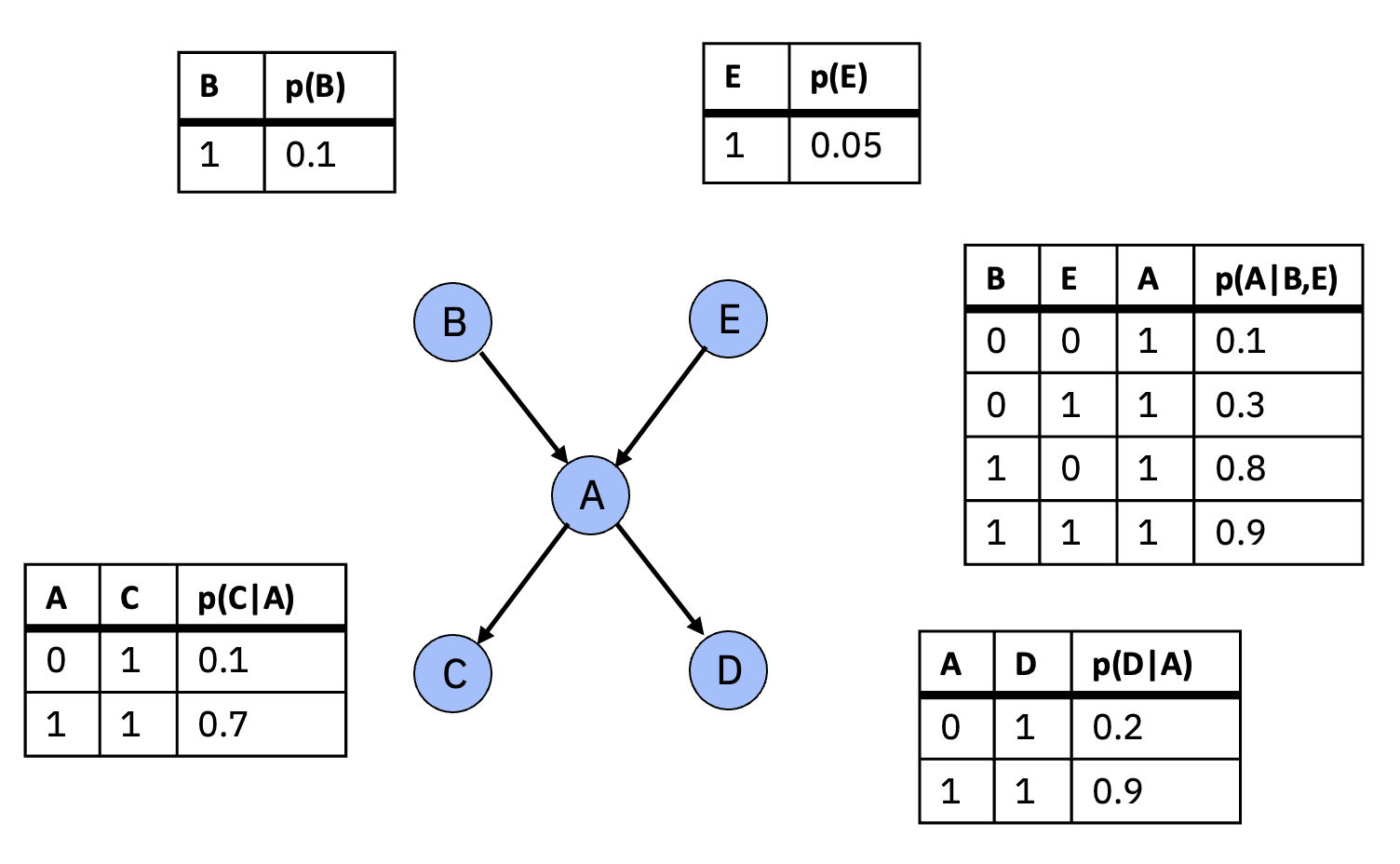}
    \caption{A simple Bayesian network.}
    \label{fig:ex-bn}
\end{figure}

\begin{example} \label{ex:burglar-and-earthquake}
    Figure \ref{fig:ex-bn} shows a simple Bayesian network defined over 5 variables $\{A,B,C,D,E\}$ having bi-valued domains $\{0,1\}$. The CPDs are shown next to the variable nodes.
\end{example}

%

\subsection{Credal Networks} \label{sec:cns}
Credal Networks \cite{COZMAN2000199, MAUA2020133} generalize Bayesian Networks -- the nodes in a Credal Network are just like the nodes in a Bayesian network, but the probabilities associated with each node are given as ranges, with associated minimum and maximum probabilities (so-called ``credal sets''. 

In a Bayesian Network one specifies a set of Conditional Probability Distributions while in a Credal Network one specifies joint credal sets.

\subsection{Logical Credal Networks} \label{sec:lcns}

A Logical Credal Network (LCN) \cite{lcn2022neurips} begins with a Credal Network and adds propositional logical sentences. Formally, an LCN $\mathcal{L}$ is defined by a set of two types of \emph{probability sentences}:
\begin{align}
  l_q \leq & p\left( q \right) \leq u_q \label{eq:syn1}\\
  l_{q \mid r} \leq & p\left( q \mid r \right) \leq u_{q \mid r} \label{eq:syn2}
\end{align}
where $q$ and $r$ can be arbitrary propositional sentences
and $0 \leq l_q \leq u_q \leq 1$, $0 \leq l_{q \mid r} \leq u_{q \mid r} \leq 1$. Each sentence in $\mathcal{L}$ is further associated with a Boolean parameter $\tau$ indicating dependence between the atoms of $q$.

An LCN represents the set of probability distributions (i.e., \emph{models}) over all interpretations that satisfy a set of constraints given explicitly by sentences (\ref{eq:syn1}) and (\ref{eq:syn2}) together with a set of implied independence constraints between the LCN's atoms. An LCN is \emph{consistent} if it has at least one model. Otherwise, it is \emph{inconsistent}.

\subsection{Logical Neural Networks}
LNNs \cite{riegel2020logical, DBLP:conf/emnlp/SenCAKRG22, DBLP:conf/aaai/SenCRG22} are a form of recurrent neural network that perform deductive logical inference on any weighted real-valued logic 
\cite{sep-logic-manyvalued, DBLP:journals/corr/abs-2008-02429}. They represent logical formulas (e.g. from a knowledge base (KB)) using collections of neurons arranged according to each formula's syntax parse tree - with a single neuron per formula symbol. Neurons specify a single activation to compute the truth value of both lower- and upper-bounds, given \textit{truth bounds} on their inputs. 
As a key differentiator from other neural networks, LNNs pass information both ``upward'' in the usual sense by the direct evaluation of their neurons' activation functions and ``downward'' from known or proved truth value bounds on each neuron's output to each of the neuron's inputs via the functional inverse of its activation function. This downward pass in effect performs various forms of deductive logical inference, including \emph{modus ponens}. 
The truth bounds representation in LNN allows for 
the recognition of inconsistencies when lower and upper bounds cross.

\begin{figure}[h]
    \centering
    \includegraphics[scale=0.67]{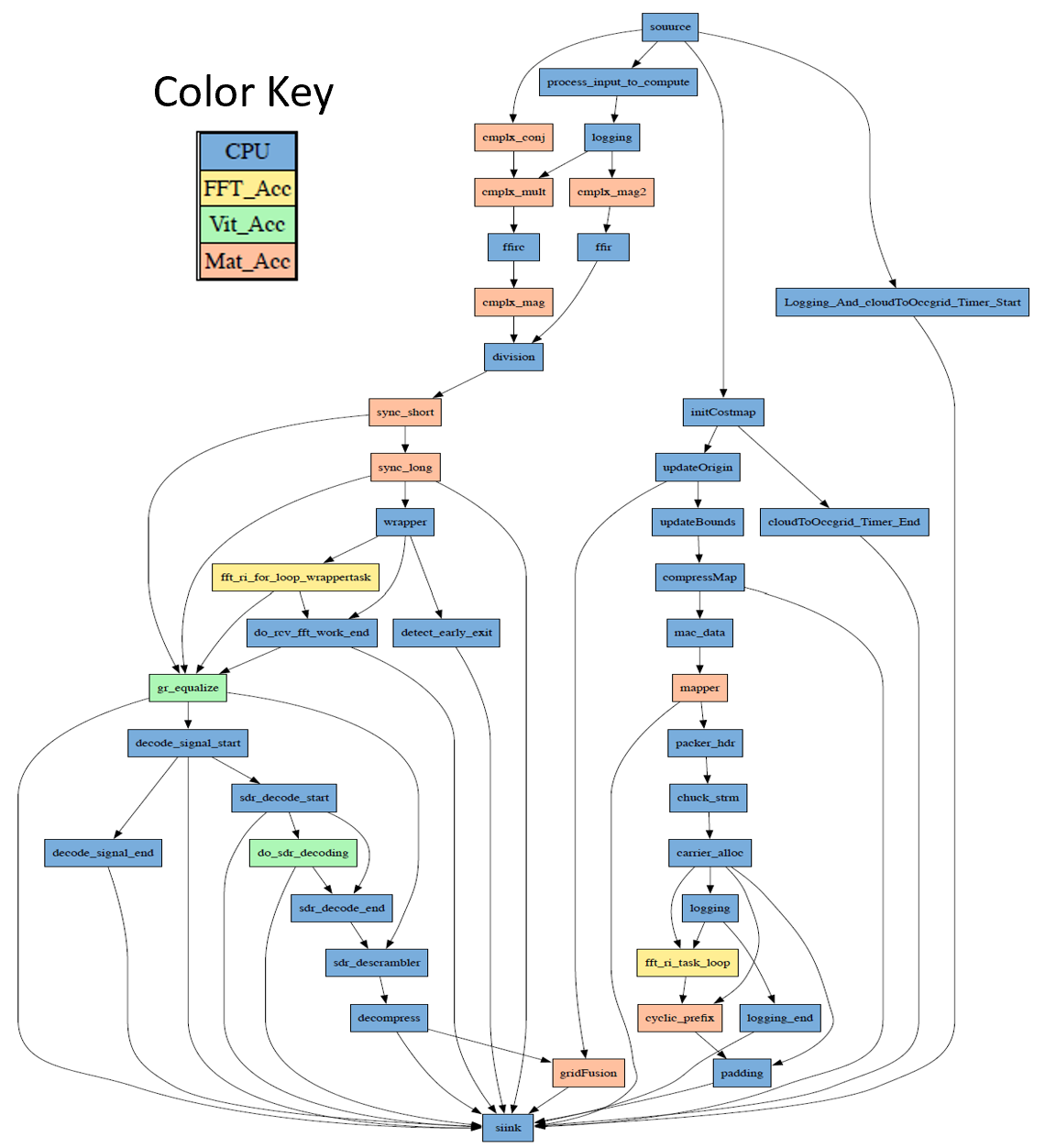}
    \caption{DAG representation of a co-operative autonomous vehicle 
    application. Colors used for the tasks indicate the PE assignment as shown in the key}
    \label{fig:eradag}
\end{figure}

\section{Upward and Downward Inference Rules for Conditionals} \label{app:up_down_condtional}

The upward and downward inference update rules for the conditional $(A|B)$ comes directly from the definition of $p(A|B)$ as $p(A \wedge B)/p(B)$. For upward inference we obtain:
\begin{eqnarray*}
    l_{(A|B)} &\mapsto& \max\bigg(l_{(A|B)}, \frac{l_{A \wedge B}}{u_B}\bigg), \\
    u_{(A|B)} &\mapsto& \min\bigg(u_{(A|B)}, \frac{u_{A \wedge B}}{l_B}\bigg).
\end{eqnarray*}

For downward, as long as $p(A|B) > 0$, we have $p(B) = p(A \wedge B) / p(A|B)$. Therefore,
\begin{eqnarray*}
\textrm{If }u_{(A|B)} > 0,~ &&
l_B \mapsto \max\bigg(l_B, \frac{l_{A \wedge B}}{u_{(A|B)}}\bigg),\\
\textrm{If }l_{(A|B)} > 0,~ &&
u_B \mapsto \min\bigg(u_B, \frac{u_{A \wedge B}}{l_{(A|B)}}\bigg).
\end{eqnarray*}
And, then, the relatively obvious:
\begin{eqnarray*}
    l_{A \land B} &\mapsto& \max(l_{A \land B}, l_B l_{(A|B)}), \\
    u_{A \land B} &\mapsto& \min(u_{A \land B}, u_B u_{(A|B)}).
\end{eqnarray*}

\section{Downward Inference Revisited} \label{app:j-update}

When we described upward and downward inference earlier we purposely oversimplified the discussion because we had not yet discussed interpolation of the \Ft Inequalities using the $J$ parameter. 
In the more complete picture, this time, say, for the case of conjunction, in the upward direction,
we assume we are given bounds $l_A, u_A$ on $A$ and $l_B, u_B$ on $B$, along with bounds $l_J,u_J$ on $J$ 
with respect to $A, B$, and we wish to compute lower and upper bounds on $A \land B$. 
It is easy to see that $p(A \land B)$ is minimized at $l_A, l_B$ and when $A$ and $B$ are maximally anti-correlated, so at $l_J$, and further that $p(A \land B)$ is maximized at the respective upper bounds. Thus, from (\ref{moduldated-frechet-and}), (\ref{p_and_-1}) -- (\ref{p_and_1}) we have:
\footnotesize
\begin{eqnarray}
l_{A \land B} &=& \frac{-l_J(1-l_J)}{2}\max(0, l_A + l_B - 1) + \label{lower_and}\\ 
&&(1+l_J)(1-l_J) l_A l_B + 
\frac{l_J(1+l_J)}{2}\min(l_A,l_B), \notag \\
u_{A \land B} &=& \frac{-u_J(1-u_J)}{2}\max(0, u_A + u_B - 1) + \label{upper_and}\\
&&(1+u_J)(1-u_J) u_A u_B + 
\frac{u_J(1+u_J)}{2}\min(u_A,u_B).\notag
\end{eqnarray}
\normalsize

The analogously more complete picture of downward inference for conjunction is a tiny bit mind-bending but runs as follows. Let us assume we are given $l_{A \land B}, u_{A \land B}$ along with $l_A, u_A$ and $l_J, u_J$. How should we update $l_B, u_B$?  We first consider $l_B$. As a function of $p(A \land B)$, keeping other quantities fixed, $p(B)$ is plainly minimized at $l_{A \land B}$. For fixed values of $p(A \land B)$, and $p(A)$, $p(B)$ is minimized when the overlap between $p(A)$ and $p(B)$ is minimized. This happens when $J$ is maximized, in other words at $u_J$. Finally, for fixed values of $p(A \land B)$ and $J$, $p(B)$ is minimized when $p(A)$ is maximized -- at $u_A$.  On the other hand, to maximize $p(B)$, all conclusions must be flipped. In other words, this maximization happens at $u_{A \land B}, l_J$ and $l_A$.

From (\ref{moduldated-frechet-and}) and (\ref{p_and_-1})-(\ref{p_and_1}), we thus have that 
\footnotesize
\begin{eqnarray}
    l_{A \wedge B} &=& \frac{-u_J(1-u_J)}{2} \max(0, u_A + l_B - 1) +\label{lower_and_dnward}\\
    && (1 + u_J)(1-u_J)u_A l_B 
     + \frac{u_J(1+u_J)}{2}\min(u_A, l_B), \notag
\end{eqnarray}
\normalsize
which we solve for $l_B$ in terms of the known quantities $l_{A \wedge B}, u_A$ and $u_J$.  And analogously:
\footnotesize
\begin{eqnarray}
    u_{A \wedge B} &=& \frac{-l_J(1-l_J)}{2} \max(0, l_A + u_B - 1)
     +\label{upper_and_dnward}\\
     && (1 + l_J)(1-l_J)l_A u_B 
     + \frac{l_J(1+l_J)}{2}\min(l_A, u_B), \notag
\end{eqnarray}
\normalsize
which we solve for $u_B$ in terms of the known quantities $u_{A \wedge B}, l_A$ and $l_J$. 

In addition to updating the bounds on $A$ or $B$ given bounds on $A \land B$, $J$ and the other of $A, B$, downward inferencing also updates the bounds on $J$ given bounds on $A \land B, A$, and $B$. 

In order to update the bounds $l_J$ and $u_J$ given known bounds on $p(A \land B), p(A)$, and $p(B)$, we again use the equations (\ref{lower_and_dnward}) and (\ref{upper_and_dnward}). In equation (\ref{lower_and_dnward}), $l_{A \land B}, u_A$ and $l_B$ are known and $u_J$ is unknown, so we get a quadratic equation with possibly two distinct roots, with $u_J$ being the greater of the values if they are unequal. Analogously, in equation (\ref{upper_and_dnward}), $u_{A \land B}, l_A$ and $u_B$ are known and $l_J$ is unknown, and we again get a quadratic equation with possibly two distinct roots, with $l_J$ being the lesser of the values if they are unequal.

\section{Real-World Application \label{app:realdag}}
Figure~\ref{fig:eradag} shows a real-world application DAG for collaborative autonomous vehicles. 
\section{PLNN Inference}
\subsection{PLNN Inference Setup \label{app:plnninfstp}}
\begin{figure*}[!ht]
    \centering
    \includegraphics[width=1.0\linewidth]{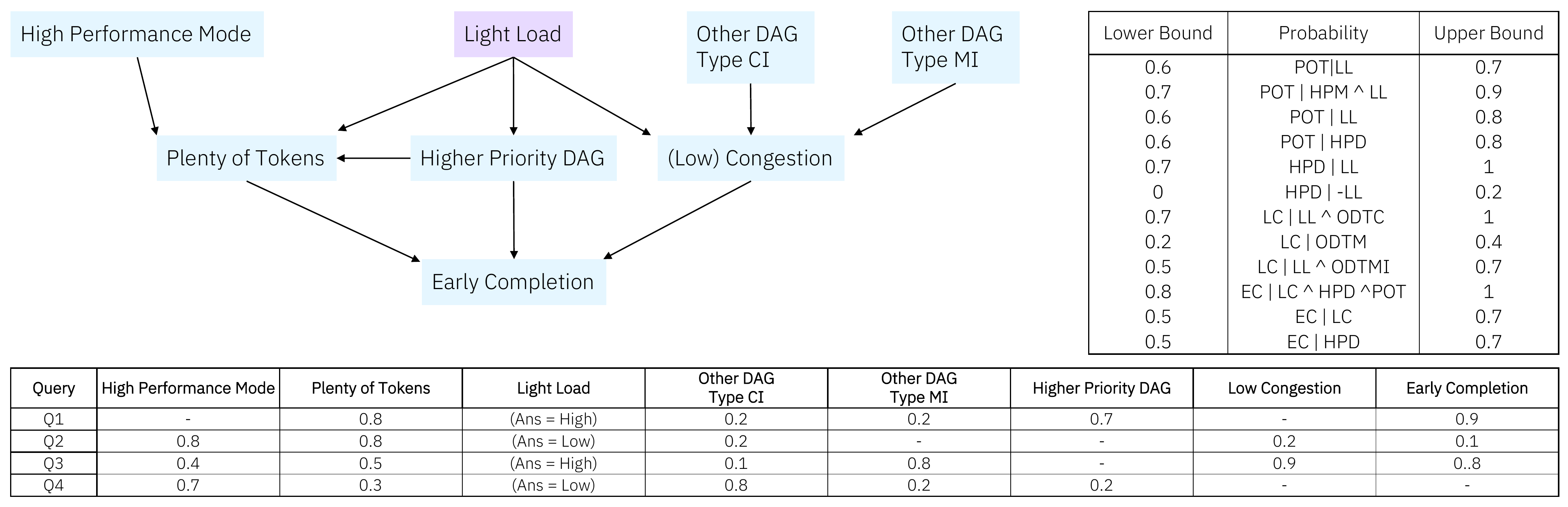}
    \caption{A DAG representing the domain knowledge of interacting variables with highlighted nodes of interest (top left), a subset of known conditional bounds (top right), 
    and a table of queries with bounds on select variables and their expected results (bottom).}
    \label{fig:plnn_exp_graph}
\end{figure*}
The purpose of PLNN inference, in the HSoC power optimization application, is for each agent to be able to predict the transient state of system variables that affect task execution times. As there are multiple tasks sharing the HSoC resources, the interactions between tasks is a primary source of stochasticity in the system. This interaction is based on rate of arrival of jobs that need processing, the workload characteristics of DAGs that are running, and system settings. These factors are used in constructing the Bayesian Network that describes cause and effect in the state of relevant system variables, and for the simplest case illustrated here, all variables only take on binary logical values (True or False) and are described by their probability of being True. Independent variables form the top layer of the Bayesian Network. These include: ‘High Performance Mode’ (HPM) which is a setting that determines the availability of power tokens, hence a True value of HPM results in a True value for the ‘Plenty of Tokens’ (POT) predicate. Other factors, such as, ‘Light Load’ (LL) in the system, or the agent’s task belonging to a ‘Higher Priority DAG’ (HPD), can similarly result in POT having a True state with a certain probability. POT and HPD, in turn, result in ‘Early Completion’ times (EC) with a probability. ‘Low Congestion’ (LC) refers to ease of network data communication between the PEs. The data sharing requirements of the DAGs in the system drives LC state. For example, if the ‘Other DAG Type CI’ (ODTCI) is True, refers to other workloads that are compute intensive (not much data communication), while ‘Other DAG Type MI’ (ODTMI) being true indicates memory intensive DAGs in the system which would predict potential network congestion and likelihood of LC being False. LC is another factor that drives Early Completion times. 

\begin{figure*}[ht]
    \centering
    \includegraphics[width=1.0\linewidth]{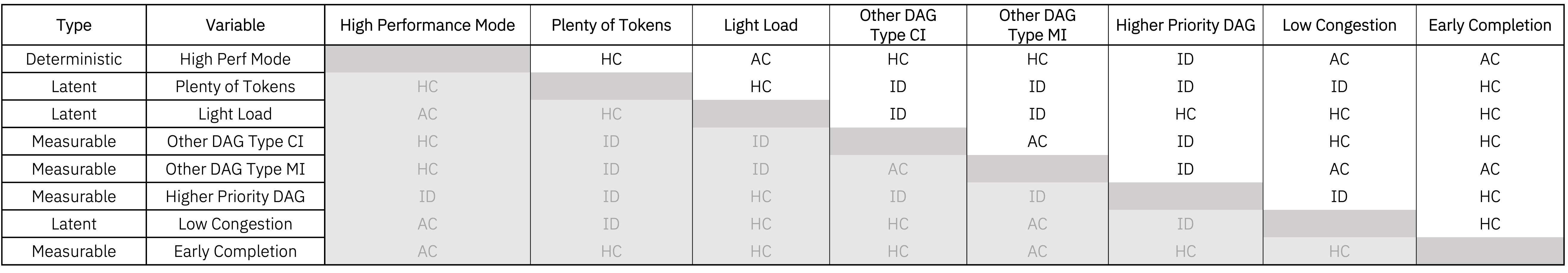}
    \caption{A table of enforceable correlations between variables for the PLNN experiment, i.e., variables can have an unknown-correlation (UC), high-correlation (HC), independent (ID) or anti-correlation (AC) behavior. The values of the correlation, in relation to the \texttt{J} parameter can be specified and provides additional flexibility. As stronger correlations are enforced, the expected behavior is for the bounds on "loose" (uncertain) nodes to be "tightend". However, introducing correlations that are too rigid can push many nodes in the graph into contradictions simultaneously.}
    \label{fig:plnn_correlations}
\end{figure*}

Thus, domain knowledge is fed into the PLNN as logic state expectations connecting predicates to be either highly correlated, anti-correlated or independent as shown in Figure~\ref{fig:plnn_correlations}. 
In addition, based on historical data, conditional probabilities can be calculated offline and provided as an additional source of information to each agent to make system state predictions. PLNN is able to handle uncertainty in these conditional probabilities arising from uncertainty in the system state, by the use of upper and lower bounds. Each agent makes its own direct observations of different measurable predicates, for example, EC, as well as aggregating EC information exchanged between siblings from the same DAG. Domain knowledge can be injected into the agent’s EC probability estimation as heuristics that include weighting factors for freshness or staleness of information timing, diversity of workloads in the information source, etc. The agent, therefore, has access to state variable information with varying levels of certainty in the course of executing runtime tasks. The agent uses PLNN inference to predict the state of latent variables, such as LL and LC, which are important for its dynamic decision-making regarding power sharing. The agent may query the LL state of the system with partial information inferred for certain predicates while missing information for other ones. The reasons for missing information could be that the agent was idle and therefore its information is stale, or conflicting information from siblings does not allow reliable probability estimation. PLNN handles these uncertainties to allow the agent to make probability bounds predictions for latent state variables.

\begin{figure*}[ht]
   \centering
   \includegraphics[width=1.0\linewidth]{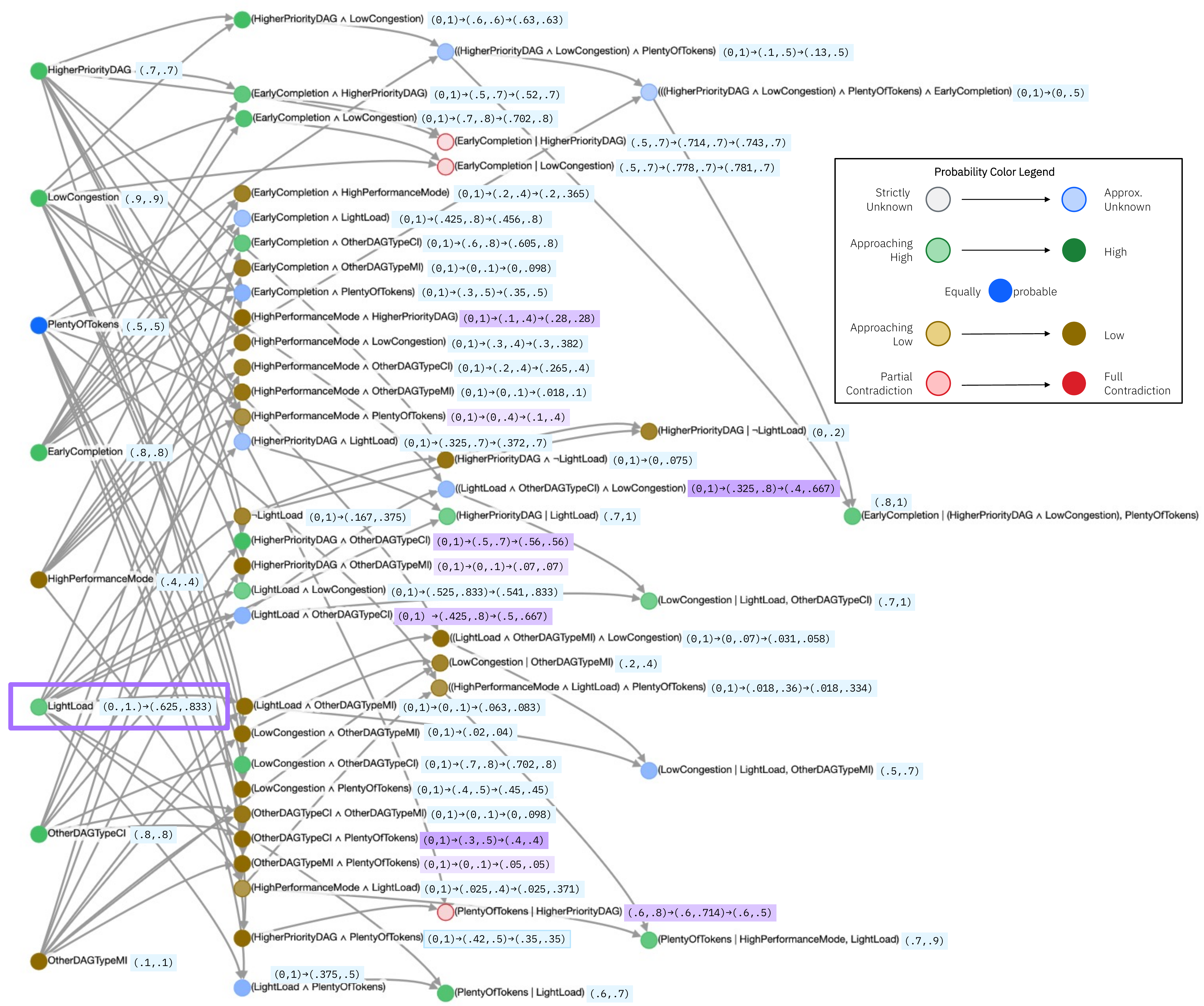}
   \caption{Visualization of the PLNN graph for Q3. 
   The bounds (in blue) represent three graph instances: (i) the instantiated graph; (ii) updates using \textit{PLNN} with loose \textit{J} values $(-1, 1)$, (iii) updates using \textit{J} values from the domain knowledge. The highlighted bounds indicate where \textit{PLNN-J} achieves meaningful updates, while omited bounds indicate no change.
   The legend arrows indicates the effect that monotonically tightening of bounds has on the probability and thus the color gradient. 
   }
   \label{fig:plnn_exp_viz}
\end{figure*}

\end{document}